\documentclass[letterpaper, 10 pt, conference]{ieeeconf}  

\IEEEoverridecommandlockouts                              

\usepackage[dvipsnames]{xcolor}
\usepackage{multirow}
\usepackage{float}
\usepackage{stfloats}

\usepackage{indentfirst}
\usepackage{caption}

\usepackage{multirow}
\usepackage{diagbox}
\usepackage{makecell} 
\usepackage{colortbl}
\usepackage{xcolor}
\usepackage{bbm}
\usepackage{subcaption}
\usepackage{textcomp}
\usepackage{stmaryrd}
\usepackage{graphicx}

\usepackage{booktabs}
\usepackage{amsmath}
\usepackage{array}
\usepackage{amsfonts}
\usepackage{adjustbox}
\usepackage{gensymb}
\usepackage{xspace}
\usepackage{cite}

\usepackage{hyperref}

\makeatletter
\DeclareRobustCommand\onedot{\futurelet\@let@token\@onedot}
\def\@onedot{\ifx\@let@token.\else.\null\fi\xspace}

\def\ie{\emph{i.e}\onedot}

\def\wrt{w.r.t\onedot} 
 
\def\etal{\emph{et al}\onedot}
\makeatother

\title{\LARGE \bf
Uncertainty-Aware Knowledge Distillation for Compact and  Efficient \\ 6DoF Pose Estimation
}

\author{Nassim {Ali Ousalah}$^{1}$, Anis Kacem$^{1}$, Enjie Ghorbel$^{1,2}$, Emmanuel Koumandakis$^{3}$ and Djamila Aouada$^{1}$
\thanks{$^{1}$SnT, University of Luxembourg, Luxembourg.}%
\thanks{$^{2}$Cristal Lab, ENSI, University of Manouba.}%
\thanks{$^{3}$Infinite Orbits, Toulouse.}%
\thanks{Corresponding author: {\tt\small nassim.aliousalah@uni.lu}}%
}

\begin{document}

\maketitle
\thispagestyle{empty}
\pagestyle{empty}

\begin{abstract}
Compact and efficient 6DoF object pose estimation is crucial in applications such as robotics, augmented reality, and space autonomous navigation systems, where lightweight models are critical for real-time accurate performance. This paper introduces a novel uncertainty-aware end-to-end Knowledge Distillation (KD) framework focused on keypoint-based 6DoF pose estimation. Keypoints predicted by a large teacher model exhibit varying levels of uncertainty that can be exploited within the distillation process to enhance the accuracy of the student model while ensuring its compactness. To this end, we propose a distillation strategy that aligns the student and teacher predictions by adjusting the knowledge transfer based on the uncertainty associated with each teacher keypoint prediction. Additionally, the proposed KD leverages this uncertainty-aware alignment of keypoints to transfer the knowledge at key locations of their respective feature maps. Experiments on the widely-used LINEMOD benchmark demonstrate the effectiveness of our method, achieving superior 6DoF object pose estimation with lightweight models compared to state-of-the-art approaches. Further validation on the SPEED+ dataset for spacecraft pose estimation highlights the robustness of our approach under diverse 6DoF pose estimation scenarios.
\end{abstract}







\section{Introduction}
\label{sec:intro}

Six Degrees of Freedom (6DoF) object pose estimation is a fundamental area in computer vision, with wide-ranging application fields such as Augmented Reality (AR), robotic manipulation, satellite docking, and space debris tracking~\cite{10043016, Lin_2024_CVPR, Nguyen_2024_CVPR, Peng_2019_CVPR, Di_2021_ICCV}. Its core objective involves estimating both the position and orientation of an object in 3D space relative to the camera coordinate system.

The task of 6DoF pose estimation is typically addressed using a two-stage pipeline: (1) establishing correspondences between the 3D model of the object and input images~\cite{Peng_2019_CVPR, Di_2021_ICCV, chen2019satellite}, and (2) computing the object's pose from these correspondences using a Perspective-n-Point (PnP) algorithm~\cite{chen2022epro, item_34721c4a90c2421c9df615a96127a4a2, li2012robust, terzakis2020consistently}. While traditional methods address the first step by relying on carefully hand-designed features~\cite{vidal20186d}, recent successful approaches have mostly shifted towards deep learning regressors that predict key primitives such as keypoints~\cite{chen2019satellite, Peng_2019_CVPR} or local predictions~\cite{hu2021wide, hu2019segmentation}, provided as input to the PnP solver.

\begin{figure}[t]
    \includegraphics[width=0.475\textwidth]{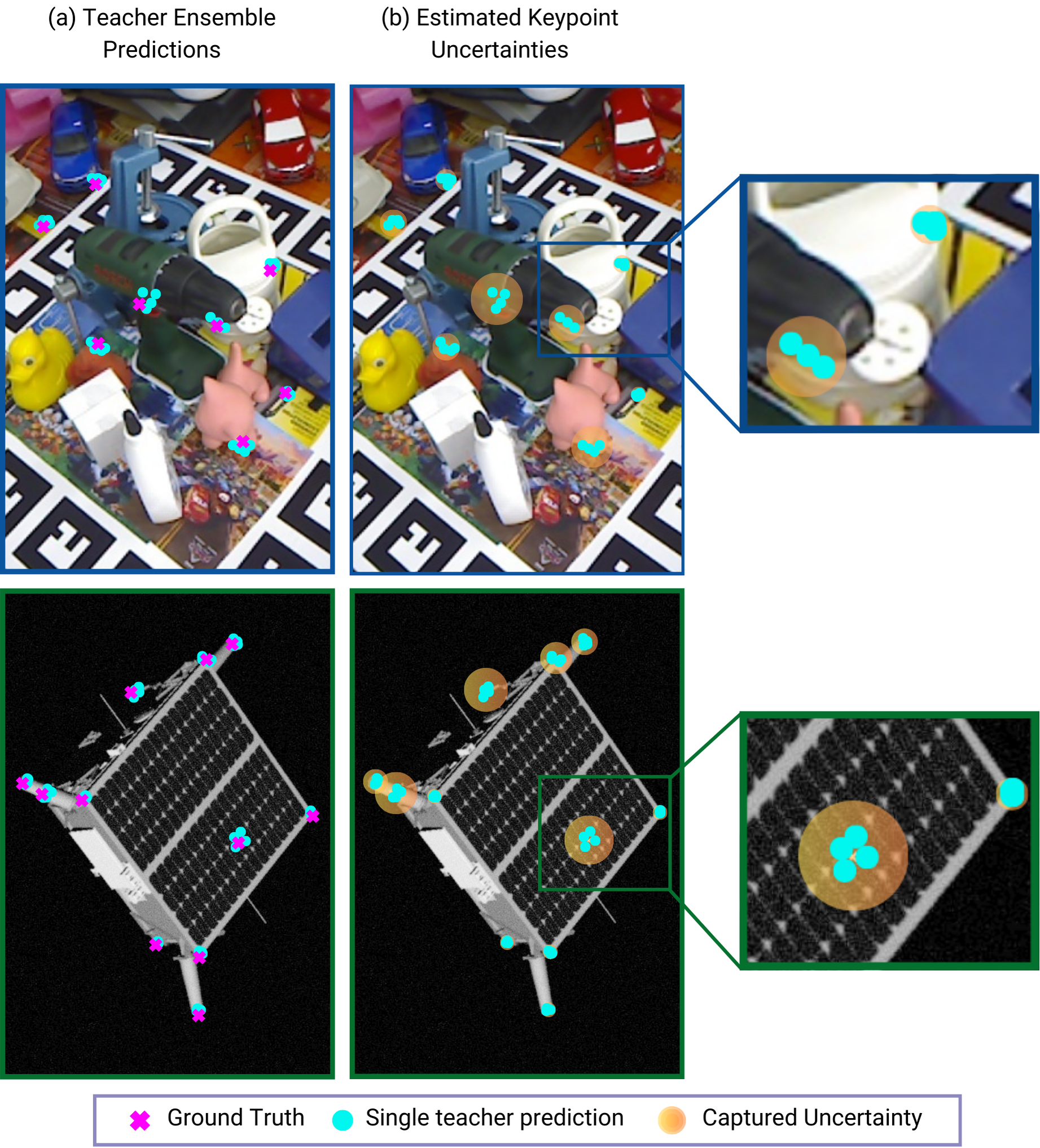}
    \caption{\textbf{Uncertainty in Teacher Predictions}. (a) Predictions from multiple teacher models show varied keypoint locations, represented by individual teacher predictions. (b) Keypoint predictions with high uncertainty values are visualized through clustered uncertainty markers.}
    \vspace{-7mm}
    \label{fig:motivation}
\end{figure}



Despite achieving great accuracy, 6DoF pose estimation methods are practically challenged by real-world constraints. A key limitation lies in their reliance on large deep neural architectures with tens of millions of parameters~\cite{park2023spnv2, kajak2021segmentation, 10.1007/978-3-031-25056-9_11}. 


To address this, Guo \etal~\cite{Guo_2023_CVPR} introduced a Knowledge Distillation (KD) strategy for efficient 6DoF pose estimation. KD aims at transferring knowledge from a deep teacher model to a more compact student model. While such an approach has been widely explored in the field of computer vision~\cite{Wang_2024_CVPR, Zhang_2024_CVPR, huangpu2024efficient}, the work of~\cite{Guo_2023_CVPR} is the first to consider it in the specific field of 6DoF pose estimation. They observed that compact student networks often struggle to predict accurate keypoints as compared to larger teacher networks. They address this by distilling knowledge from the teacher to the student at the \textit{prediction-level} of the keypoints through an Optimal Transport (OT) approach~\cite{villani2009optimal}. 
Additionally, the \textit{prediction-level} KD was naively combined with an off-the-shelf \textit{feature-level} KD method~\cite{zhang2021improve} in an attempt to further enhance the accuracy of the student model.

While KD stands as an appealing approach for achieving efficient 6DoF pose estimation~\cite{guan2022hrpose, Guo_2023_CVPR}, we identify key limitations at two levels in existing methods: (1) \textit{prediction-level -} most works assume that the teacher network predicts all keypoints with equal \textit{confidence}. However, as shown in Figure~\ref{fig:motivation}-b, the keypoint predictions from the teacher often exhibit varying levels of \textit{uncertainty}. Current methods~\cite{Guo_2023_CVPR, guan2022hrpose} treat all predicted keypoints equally, ignoring this uncertainty. As a result, irrelevant knowledge may be transferred from the teacher, causing the introduction of a bias in the student model towards less reliable teacher predictions. (2) \textit{feature-level -} KD is addressed at the \textit{feature-level} independently from the \textit{prediction-level}, which can lead to inconsistencies when aligning the features and predictions between the teacher and student networks.

In this paper, we propose a novel KD method at both the\textit{ prediction-level} and the \textit{feature-level} for efficient keypoint prediction in 6DoF object pose estimation, addressing the identified limitations. Our \textit{prediction-level} KD approach, referred to as Uncertainty-Aware Knowledge Distillation (UAKD), leverages the uncertainties of keypoints predicted by the teacher model to guide the alignment between the teacher and student prediction distributions. The influence of keypoints with higher uncertainty in the teacher model is reduced in the KD process. This is achieved through an uncertainty-aware OT approach, where uncertainty values are integrated into the alignment strategy using an unbalanced Sinkhorn algorithm~\cite{sejourne2019sinkhorn}. As shown in Figure~\ref{fig:motivation}-a, we rely on a teacher ensembling method~\cite{lakshminarayanan2017simple, wursthorn2024uncertainty} to estimate the keypoint uncertainties, as 6DoF pose estimation techniques do not necessarily provide explicitly this information.  

In addition, we extend our KD framework to operate at the \textit{feature-level}. The proposed KD mechanism is termed Prediction-related Feature Knowledge Distillation (PFKD). In particular, the predicted keypoints are traced back to their corresponding receptive fields in the feature maps of both the teacher and student models. By applying the OT solution from the \textit{prediction-level} to the feature maps, we establish correspondences between relevant regions in the teacher and student feature maps. This enables distillation at key locations in the feature space. By leveraging the \textit{prediction-level} OT solution in the \textit{feature-level} KD, we develop a comprehensive end-to-end uncertainty-aware KD approach that leverages keypoints along with their related feature representations from the teacher model.
Experiments conducted on two publicly available datasets, namely LINEMOD~\cite{10.1007/978-3-642-37331-2_42} for regular object pose estimation and SPEED+~\cite{park2022speed+} for satellite pose estimation using two different backbones~\cite{hu2021wide,park2023spnv2}, demonstrate that our method outperforms existing KD approaches for efficient 6DoF pose estimation~\cite{guan2022hrpose, Guo_2023_CVPR}.

\noindent Our main contributions can be summarized as follows. 

    
    

\begin{itemize}
    \item An Uncertainty-Aware \textit{prediction-level} KD (UAKD) for efficient keypoint prediction in 6DoF pose estimation. 
    \item A \textit{feature-level} KD approach (PFKD) that is consistent with the \textit{prediction-level} KD by leveraging the \textit{prediction-level} OT solution to establish correspondences between feature map regions and keypoint predictions in both the teacher and student models.
    \item Extensive experiments on two well-known datasets with two state-of-the-art pose estimation models~\cite{hu2021wide,park2023spnv2}.
\end{itemize}
\section{Related Work}
\label{sec:relatedwork}
In this section, we start by presenting the state-of-the-art on efficient 6DoF pose estimation, then review related works on the field of KD. 

\noindent \textbf{Efficient 6DoF Pose Estimation.} With the growing demand for real-time applications, research on pose estimation has increasingly emphasized the need for accurate yet computationally efficient solutions~\cite{Kehl_2017_ICCV,Tekin_2018_CVPR,bukschat2020efficientpose,guan2022hrpose,Guo_2023_CVPR}. Efforts have been therefore dedicated for introducing pose estimation methods~\cite{Kehl_2017_ICCV, Tekin_2018_CVPR,tan2020efficientdet} that employ computationally friendly backbones such as SSD~\cite{liu2016ssd} and EfficientDet~\cite{tan2020efficientdet}.
Nevertheless, these approaches predominantly prioritize the computational cost over other deployment challenges such as memory constraints. As a result, while speed is a crucial aspect, taking into account the number of parameters used in a model is also essential for reaching real-world standards in efficiency. To this end, Guan \etal~\cite{guan2022hrpose} were the first to use knowledge distillation to train an efficient 6DoF pose estimation model based on the HRNetV2-W18 backbone~\cite{wang2020deep}. Additionally, Guo \etal~\cite{Guo_2023_CVPR} proposed a KD approach by aligning keypoint predictions from a highly accurate teacher model to a less performing and compact one using optimal transport~\cite{villani2009optimal}. Despite being promising, current methods treat all keypoints equivalently while having different levels of uncertainty; therefore hindering the distillation process. Moreover, they decouple the KD at the feature and the prediction levels, leading to inconsistent knowledge transfer.  


\noindent \textbf{Knowledge Distillation (KD).} KD has proved to be an effective way to transfer knowledge from a well-performing deeper teacher network to a more compact student one,  with the aim of improving the accuracy of the student model~\cite{hinton2015distilling,Wang_2024_CVPR,Son_2021_ICCV,Jin_2023_CVPR,Wang_2023_CVPR,svdKD}. KD was first proposed by Hinton~\etal~\cite{hinton2015distilling} where the authors tried to leverage the last layer outputs of the teacher network. Inspired by this work, many KD variants have been derived by considering different strategies~\cite{Wang_2024_CVPR,Son_2021_ICCV,ji2021show,svdKD,Passban_Wu_Rezagholizadeh_Liu_2021,Yang_2022_CVPR} such as the use of intermediate features from the head~\cite{Wang_2024_CVPR} or multiple teacher assistants~\cite{Son_2021_ICCV}. Nevertheless, most of these approaches do not transfer the knowledge coherently at the \textit{prediction-level} and the \textit{feature-level}, which may cause inconsistencies. Only few methods based on attention~\cite{wang2020deep, Passban_Wu_Rezagholizadeh_Liu_2021, ji2021show} tried to address this issue on different computer vision tasks; however, disregarding the specific field of 6DoF pose estimation. Moreover, none of these approaches have considered the uncertainty aspect, which is a core component in this paper. This highlights the need to develop a feature distillation mechanism compatible with an uncertainty-aware \textit{prediction-level} KD.

\section{Problem Statement}
\label{sec:problemstatement}

Estimating the 6DoF pose of a given object from a sample \( \mathbf{I} \in \mathcal{I} \) belonging to the space of RGB images $\mathcal{I}$  involves finding a function \( \mathbf{\Phi} \) such that,
\begin{equation}
    \begin{array}{ccccc}
    \renewcommand{\arraystretch}{0.2}
        \mathbf{\Phi} ~~  : & \mathcal{I} & \longrightarrow & \mathcal{SO}(3) \times \mathbb{R}^{3} \ , \\
         & \mathbf I & \longmapsto & \mathbf{\Phi}(\mathbf I) = (\mathbf R, \mathbf t) \ ,
    \end{array}
\end{equation}
\noindent where \( (\mathbf{R}, \mathbf{t}) \) represents the object pose formed by a rotation matrix \( \mathbf{R}~\in~\mathcal{SO}(3) \) and a translation vector  \( \mathbf{t} \in \mathbb{R}^3 \)  \wrt the camera coordinate system. \( \mathcal{SO}(3) \) denotes the special orthogonal group in dimension 3. 

In a two-stage neural 6DoF pose estimation method~\cite{Peng_2019_CVPR, Di_2021_ICCV, chen2019satellite}, the first stage relies on a neural network $\mathbf \Phi_k$ that detects from an input \( \mathbf{I} \), a total of $N$ 2D keypoints $ \mathbf K~=~(\mathbf k_i)_{i\in [[1,N]]}$, where 
 $\mathbf k_i~\in~\mathbb{R}^{ 2}$ and $\mathbf K~\in~\mathbb{R}^{N\times2}$. 
Using the predicted 2D keypoints \( \mathbf K \) and known keypoints \( \mathbf  K_{3D} \in \mathbb{R}^{N \times 3} \) from a 3D object model, the pose parameters can be estimated using a PnP solver~\cite{chen2022epro, li2012robust}, \ie,~$(\mathbf{R},\mathbf{t})~=~\mathbf{PnP}(\mathbf{\Phi}_k(\mathbf I), \mathbf K_{3D})$. The performance of two-stage 6DoF pose estimation methods heavily depends on the accuracy of the predicted keypoints $\mathbf K$. Smaller keypoint prediction models tend to have a higher predicted pose error compared to larger ones~\cite{Guo_2023_CVPR}. 
Knowledge Distillation~(KD) addresses this challenge by considering a trained well-performing large teacher model $\mathbf{\Phi}_k^{T}$, parametrized by $\theta_T$, and transferring its knowledge when training a smaller student model $\mathbf{\Phi}_k^{S}$, parametrized by $\theta_S$, for the same task, \ie, keypoint prediction. Here, it is important to note that $|\theta_S| < | \theta_T|$, where $|.|$ refers to the cardinality. The student model is then tasked with two learning objectives at the same time, namely, the minimization of the error between the predictions and the ground-truth, while matching its feature activations and/or predictions with those of the teacher. The parameters of the student are optimized as follows, 
\begin{equation} \label{eq:pbstatement} 
    \underset{\theta_S}{\arg \min} \gamma_{\text{kpt}} \mathcal{L}_{\text{kpt}}(\mathbf  K^S, \mathbf  K_{gt}) + \gamma_{\text{distill}}  \mathcal{L}_{\text{distill}}(\mathbf{\Phi}^S_k, \mathbf{\Phi}^T_k) \ ,
\end{equation}
\noindent where $\mathcal{L}_{\text{kpt}}(\mathbf K^S, \mathbf K_{gt})$ measures the error between the predicted student keypoints $\mathbf K^S$ and the corresponding ground-truth keypoints $\mathbf K_{gt}$, $\mathcal{L}_{\text{distill}}(\mathbf{\Phi}^S_k, \mathbf{\Phi}^T_k)$ quantifies the discrepancy between the feature activations and/or the predicted keypoints of the student~$\mathbf{\Phi}^S_k$ and the teacher~$\mathbf{\Phi}^T_k$. The parameters $\gamma_{\text{kpt}}$ and $\gamma_{\text{distill}}$ balance the two terms in~(\ref{eq:pbstatement}).

Our goal is to  define the knowledge transfer process for keypoint prediction in 6DoF pose estimation $\mathcal{L}_{\text{distill}}$ as an end-to-end distillation process that operates at two levels. Hence, we formulate it as follows, 
\begin{align}
    \mathcal{L}_{\text{distill}}(\mathbf{\Phi}^S_k, \mathbf{\Phi}^T_k) &=  
    \gamma_p \underbrace{\mathcal{L}_{\text{pred}}(\mathbf K^S, \mathbf K^T)}_{\text{Prediction-level}} \notag\\
    &\quad +  
    \gamma_f \underbrace{\mathcal{L}_{\text{feat}}(\mathbf{R}^S, \mathbf{R}^T)}_{\text{Feature-level}} \ .
\end{align}
\noindent  At the \textit{prediction-level}, we propose to align the student predicted keypoints, $\mathbf K^S$, with those of the teacher, $\mathbf K^T$, while accounting for potential uncertainties in the teacher predictions. At the \textit{feature-level}, our goal is to ensure the alignment between student and teacher feature activations $\mathbf{F}^S$  and $\mathbf{F}^T$, at key locations $\mathbf{R}^S~=~f(\mathbf{F}^S, \mathbf{K}^S )$ and $\mathbf{R}^T~=~f(\mathbf{F}^T, \mathbf{K}^T)$, respectively. These key regions are retrieved through a function $f(.,.)$ driven by keypoint predictions for ensuring a consistent end-to-end process between the \textit{feature-level} and \textit{prediction-level} alignments. 

\section{Proposed Approach}
\label{sec:metodology}

\begin{figure*}[t]
    \centering
    \includegraphics[width=0.85\textwidth]{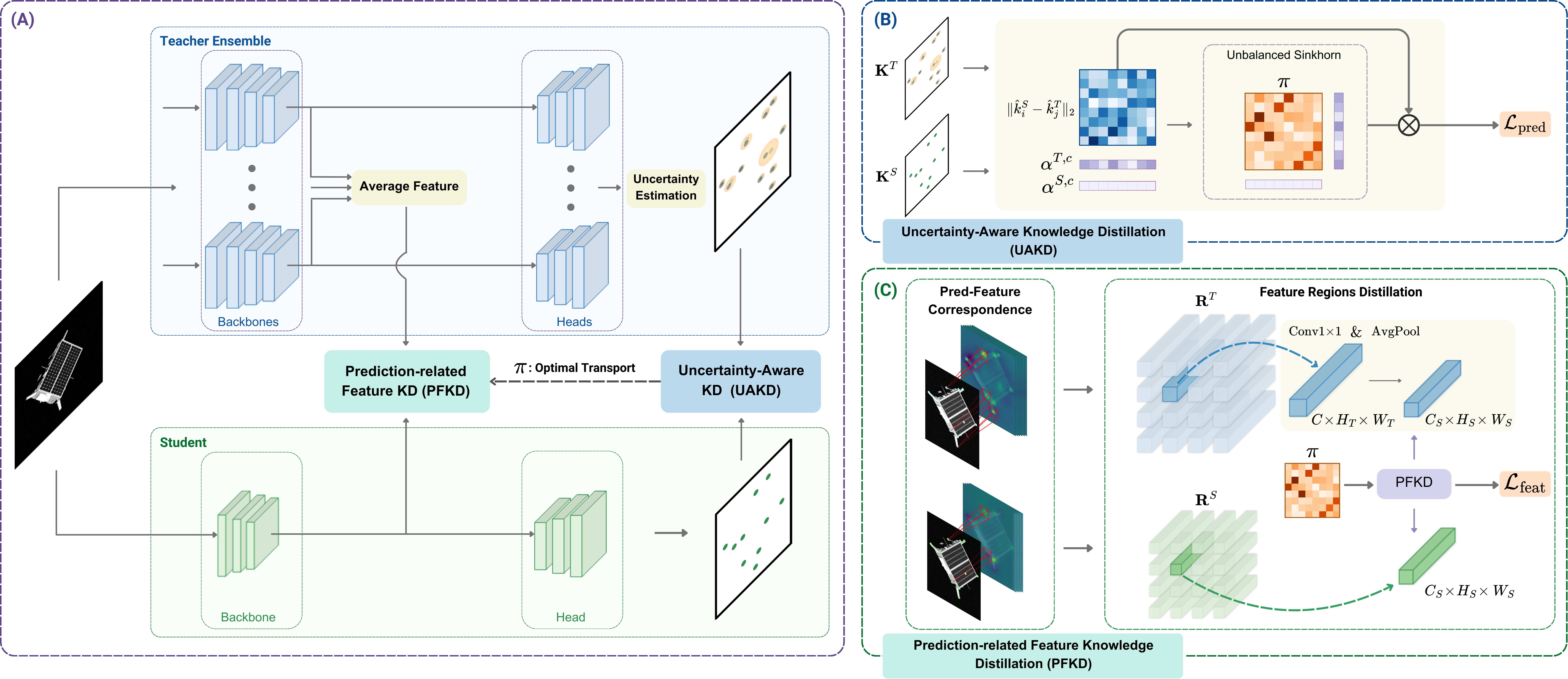}
    \caption{\textbf{Overview of the Proposed Knowledge Distillation (KD) Framework} (Best viewed in color). \textbf{(A) General overview}:~Keypoints predicted by the ensemble of teachers are used to estimate uncertainties and are subsequently processed by UAKD, while averaged feature maps are directed to PFKD, guided by the OT plan $\boldsymbol{\pi}$. \textbf{(B) UAKD Module}:~Keypoints, along with corresponding confidence scores $\boldsymbol{\alpha^{T,c}}$ and $\boldsymbol{\alpha^{S,c}}$, are aligned using an unbalanced Sinkhorn algorithm~\cite{sejourne2023unbalanced}, where $\bigotimes$ represent the tensor product. \textbf{(C) PFKD Module}:~Predicted keypoints are mapped back to their respective regions in the feature maps and are aligned consistently according to the OT plan $\boldsymbol{\pi}$.}
    \label{fig:pipeline}
    \vspace{-5mm}
\end{figure*}

To address the problem outlined in Section~\ref{sec:problemstatement}, we propose a Knowledge Distillation (KD) framework for efficient keypoint prediction for 6DoF pose estimation, integrating two complementary KD methods, namely, UAKD which operates at the \textit{prediction-level} and PFKD which acts at the \textit{feature-level}. In particular, UAKD leverages the prediction uncertainties of the teacher model for performing an Optimal Transport (OT)~\cite{villani2009optimal} alignment between the teacher and the student prediction distributions. On the other hand, PFKD relies on the correspondences established between teacher and student keypoints at the \textit{prediction-level} to guide the \textit{feature-level} distillation process, focusing on activated feature regions. Figure~\ref{fig:pipeline}-(A) illustrates the two-level overall proposed KD framework.

\subsection{Uncertainty-Aware Prediction-Level KD}
\label{sec:unc_kd}

Let $\mathbf K^{T}~=~(\hat{\mathbf  k}^T_i)_{i\in [[1,N]]}$ and $\mathbf  K^{S}~=~(\hat{\mathbf  k}^S_i)_{i\in [[1,M]]}$ be the teacher and the student predicted keypoints, respectively. To transfer the knowledge from teacher to student keypoints with different cardinalities ($N \neq M$), the keypoint alignment process can be seen as an alignment of two distributions, and therefore formulated as an unbalanced OT problem~\cite{Guo_2023_CVPR, sejourne2023unbalanced}, where each keypoint is weighted. Formally, this is equivalent to finding the optimal transportation plan $\boldsymbol \pi \in \mathbb{R}^{N \times M}$ that minimizes the overall alignment defined as,
\begin{equation}
   \begin{array}{cc}
    \mathcal{L}_{pred} = \underset{\boldsymbol \pi}{\min} \sum_{i=1}^{M} \sum_{j=1}^{N} \pi_{ij} \|\hat{\mathbf k}^S_i - \hat{\mathbf k}^T_j \|_2 \ , \\ [10pt]

    \text{s.t.} \quad \forall i, \sum_{j=1}^{N} \pi_{ij} = \alpha^{S}_i, \quad \forall j, \sum_{i=1}^{M} \pi_{ij} = \alpha^{T}_j \ ,
    
    \end{array}
   \label{eq:OT}
\end{equation}
\noindent where $\boldsymbol{\alpha}^{T}=(\alpha^{T}_1, \alpha^{T}_2, \cdots, \alpha^{T}_N)$ and $\boldsymbol{\alpha}^{S}~=~(\alpha^{S}_1, \alpha^{S}_2, \cdots, \alpha^{S}_M)$ denote the weights assigned to the teacher and student keypoint predictions, respectively. The solution to this OT problem can be efficiently found using an unbalanced Sinkhorn algorithm~\cite{NIPS2013_af21d0c9,sejourne2019sinkhorn}.

While Guo \etal adopt such an OT-based KD method~\cite{Guo_2023_CVPR}, they only consider the existence probabilities of predicted keypoints that are provided by some keypoint regressors~\cite{hu2021wide}. In this work, we posit that incorporating the uncertainties of the teacher keypoint predictions within the KD process can improve the performance of the student model. To that end, we define the following \textit{confidence} weights for the teacher and the student models, 
\begin{equation}
    \boldsymbol \alpha^{T,c} = \mathbf{1}_{N} - \mathbf{u}, \quad \boldsymbol \alpha^{S,c} = \frac{1}{M} \cdot \mathbf{1}_{M} \ ,
    \label{eq:confidence}
\end{equation}
\noindent where $\mathbf{1}_{N}$ and $\mathbf{1}_{M}$ denote $N$ and $M$-dimensional all-ones vectors, respectively, and $\mathbf{u} = (u_1, u_2, \cdots, u_N) \in [0,1]^{N}$ is a vector containing the uncertainty scores for each predicted keypoint $\hat{\mathbf k}^T_i$. 
For models that provide keypoint existence probabilities~\cite{hu2021wide}, these confidence weights can be integrated as follows, 
\begin{equation} \label{eq:lambda}
    \boldsymbol{\alpha}^{T} = \lambda \times \boldsymbol{\alpha}^{T,c} + (1 - \lambda) \times \boldsymbol{\alpha}^{T,e} \ ,
\end{equation}
\noindent where $\lambda$ denotes a modulating factor and $\boldsymbol{\alpha}^{T,e}$ are the existence probabilities of the teacher's keypoints. We chose summation over multiplication due to superior empirical performances. Figure~\ref{fig:pipeline}-(B) depicts the proposed UAKD. 
\noindent \textbf{Teacher Keypoint Uncertainty Estimation}. Here, we describe the estimation of the teacher keypoint uncertainties $\mathbf{u}$ used to define the confidence weights $\boldsymbol{\alpha}^{T,c}$ in~(\ref{eq:confidence}). 
We focus on estimating the teacher \textit{epistemic} uncertainty only, which typically arises from the model's parameters themselves~\cite{hullermeier2021aleatoric}, preventing the distillation from being influenced by the noise inherent in the data, i.e., \textit{aleatoric} uncertainty. For this end, we employ the deep ensembling method~\cite{lakshminarayanan2017simple} which approximates the posterior distribution of the teacher model's weights $ p(\theta_{T} | \mathcal{I}) $ in a similar way to Monte Carlo sampling~\cite{SHAPIRO2003353}. We represent the ensemble of trained teacher weights as $ \Theta = \{\theta_{T_{j}}\}_{j=1}^{E}$. These teacher models are similarly trained but with different weight initializations. At inference, for each input \( \mathbf{I} \in \mathcal{I} \), the trained ensemble produces $E$ sets of 2D keypoint predictions, $\mathbf K_{\Theta}^{T} = \{ (\hat{ \mathbf k}_{i}^{T_{j}})_{i\in [[1,N]]}\}_{j\in [[1,E]]}$
\noindent where $ \hat{\mathbf k}_{i}^{T_{j}} \in \mathbb{R}^{2} $ represents the $i$-th keypoint predicted by the model with the weights $\theta_{T_{j}}$. 
We then aggregate the predictions of the teachers for each keypoint $i$ by estimating its per-coordinate mean $(\mu_{x,\hat{k}_i},\mu_{y,\hat{k}_i})$ and variance $(\sigma_{x,\hat{k}_i}^2,\sigma_{y,\hat{k}_i}^2)$. The collection of mean keypoints are then used as representative of the teacher ensemble predictions $\mathbf{K}^{T}$. 
For simplicity, we consider the \( x \)- and \( y \)- axes independent, Therefore, the total variance of keypoint $i$ is given by $\sigma_{\hat{k}_i}^2 = \sigma_{x, \hat{k}_i}^2 + \sigma_{y, \hat{k}_i}^2$.
We then form $\boldsymbol{\sigma}^{T}~=~(\sigma_{\hat{k}_i}^2)_{i\in [[1,N]]}$ as the per-keypoint variances of the teacher ensemble. These variance values are further mapped to uncertainty scores in $[0,1]^{N}$ using a $\texttt{tanh}$ function to obtain $\mathbf{u}~=~\texttt{tanh}(\boldsymbol{\sigma}^{T})$. Our experiments demonstrated that an ensemble of just 4 to 6 models is sufficient to closely approximate the true variance, \ie, the epistemic uncertainty, of the teacher model.

Note that in the above, we assume that all teacher models predict the same number of keypoints. However, this may not always hold. In such cases, a majority voting is used to identify the keypoints for which uncertainties will be estimated. The remaining are assigned an uncertainty of 1.



\subsection{Prediction-related Feature-per-Keypoint KD}
\label{sec:unc_fkp_kd}

We now describe the proposed \textit{Prediction-related Feature-per-Keypoint} Knowledge Distillation (PFKD). PFKD leverages the predicted keypoints and the transportation plan obtained at the \textit{prediction-level} KD (Section~\ref{sec:unc_kd}) to distill the knowledge at key locations of the feature space.


Let $\mathbf{F}^T \in \mathbb{R}^{C \times H \times W}$ denote the feature maps extracted from the teacher backbone network, where $C$, $H$, and $W$ are the number of channels, height, and width of the feature maps, respectively. We start by retrieving the keypoint coordinates predicted by the teacher model, represented as $\mathbf{K}^{T}~=(\hat{ \mathbf k}_{i}^{T})_{i\in [[1,N]]}$.  A mapping is then established between each keypoint and the feature maps, enabling the identification of keypoint locations within the feature space. Indeed, the predicted keypoints are obtained through a series of convolutions applied on the feature maps that conserve spatial awareness, making it possible to map each keypoint to its corresponding feature location. Thus, we associate each predicted keypoint $\hat{\mathbf k}_{i}^{T}$ defined by its 2D position $\hat{\mathbf k}_{i}^{T}~=~(\hat{x}_{i}^{T}, \hat{y}_{i}^{T})$  with a feature region $\mathbf R_{i}^{T} \in \mathbb{R}^{C\times H_{T} \times W_{T}}$ centered at $c_{i}$ and with spatial dimensions $H_{T}$ and $W_{T}$ that are defined as described below,
\begin{equation}
   \left\{
   \begin{array}{ll}
       c_{i} = (\lfloor \delta \times \hat{x}_{i}^T \rceil, \lfloor \delta \times \hat{y}_{i}^T \rceil), \\ [8pt]
       H_{T} = W_{T} = 1 + \sum_{i=1}^{L} \left( (\kappa_i - 1) \prod_{j=1}^{i-1} s_j \right),
    \end{array}
    \right.
\end{equation}

\noindent where $L$ denotes the number of convolutional layers in the keypoint prediction head, $\delta$ is the scaling factor relating the feature map size to the output size, $\lfloor \cdot \rceil$ the rounding operation, and $\kappa_i$ and $s_i$ are the kernel size and stride of the $i$-th convolutional layer. We then obtain a set~$\mathbf{R}^{T}~=~\{ \mathbf R_{1}^T, \mathbf R_{2}^T, \dots, \mathbf R_{N}^T\}$ that represents the feature regions that contributed to the prediction of each keypoint. Applying a similar procedure with the student model, we obtain $\mathbf{R}^{S} = \{ \mathbf R_{1}^S, \mathbf R_{2}^S, \dots, \mathbf R_{M}^S\}$, where $\mathbf R_{i}^S~\in~\mathbb{R}^{C_{S} \times H_{S} \times W_{S}}$. If necessary, we apply a $1 \times 1$ convolution on the feature regions extracted from the teacher model to adjust the channel dimensions and an average pooling operation to modify the spatial dimensions. This ensures that the elements of $\mathbf{R}^{T}$ are transformed to $\mathbf R_{i}^T~\in~\mathbb{R}^{C_{S}, H_{S}, W_{S}}$.

Given $\mathbf{R}^{T}$ and $\mathbf{R}^{S}$, the proposed approach is turned into aligning the sets of teacher and student per-keypoint feature regions. To match each region from $\mathbf{R}^{T}$ to its corresponding one in $\mathbf{R}^{S}$, we propose to use the optimal transport plan found in~(\ref{eq:OT}). 
Thus, the proposed PFKD loss can be formulated as follows,
\begin{equation}
\resizebox{.9\hsize}{!}{$
    \begin{split}
    & \mathcal{L}_{\text{feat}}(\mathbf{R}^{T}, \mathbf{R}^{S}, \boldsymbol \pi) = \frac{1}{N \cdot M} \sum_{i=1}^{N} \sum_{j=1}^{M} \Biggl[  \\ &\quad \pi_{i,j} \sum_{d_1=1}^{C_S} \sum_{d_2=1}^{W_S} \sum_{d_3=1}^{H_S} \frac{\left( \mathbf R_{i}^T [d_1, d_2, d_3] - \mathbf R_{j}^S [d_1, d_2, d_3] \right)^2}{C_S \cdot H_S \cdot W_S} \Biggr] \ .
    \end{split}
    $}
    \label{eq:fkp}
\end{equation}



\noindent 
It is important to mention that by re-using the transportation plan of the uncertainty-aware \textit{prediction-level} KD, the proposed PFKD also incorporates the uncertainty information of the teacher. Note that the PFKD process explained above considers a single teacher model. However, our approach uses an ensemble of teachers as mentioned in Section~\ref{sec:unc_kd}. Thus, we aggregate the different feature regions extracted from different teacher models by averaging them. As a result, the set of the teacher feature regions becomes $\mathbf{R}^{T} = \{\frac{1}{E}\sum_{j=1}^{E}\mathbf R_{i}^{T_{j}}\}^{N}_{i=1}$. Figure~\ref{fig:pipeline}-(C) depicts the proposed PFKD.

Finally, the uncertainty-aware \textit{prediction-level} and \textit{feature-level} KD are together in the overall training. This is achieved by plugging their respective losses defined in~(\ref{eq:fkp}) and~(\ref{eq:OT}) into~(\ref{eq:pbstatement_distill}), where $\gamma_p$, $\gamma_f \in \mathbb{R}$ are the two modulating factors for each loss. We set $\gamma_p~=~5$ and $\gamma_f~=~0.1$ in our experiments, which leads to the best results.
\section{Experiments}
\label{sec:experiments}

In this section, we first describe our experimental setup. Then, we demonstrate the effectiveness of the proposed KD method on a well-known benchmark for generic object pose estimation, namely,  the LINEMOD dataset~\cite{10.1007/978-3-642-37331-2_42}. Furthermore, we consider the specific scenario of spacecraft pose estimation given the relevance of efficient models in the context of space applications. In particular, we validate our approach on the spacecraft pose estimation dataset SPEED+~\cite {park2022speed+}. Additional experiments are finally conducted to validate the proposed approach. 

\subsection{Experimental Setup}

\noindent\textbf{Models.} We evaluate our approach using two distinct two-stage pose estimation networks, namely, WDRNet~\cite{hu2021wide} and SPNv2~\cite{park2023spnv2}. WDRNet uses a DarkNet-53~\cite{redmon2018yolov3} backbone coupled with a feature pyramid network~\cite{lin2017feature} to predict 2D keypoint locations across multiple scales. It then employs a sampling strategy that enables feature vectors at each level to contribute probabilistically to the predictions. This results in estimating 2D keypoint clusters for each 3D bounding box corner, thus making it a local prediction-based keypoint detector. On the other hand, SPNv2 is a multi-task CNN that uses a shared multi-scale feature encoder built on an EfficientDet~\cite{tan2020efficientdet} backbone and BiFPN layers. It consists of multiple prediction heads for various tasks. In our work, we focus on the 2D keypoint prediction head that is trained to predict 2D heatmaps associated with each 2D keypoint on the object of interest. In our experiments, for the teacher models, we use WDRnet with Darknet-53 and the $\phi$~=~6 version of SPNv2 based on the EfficientDet-D6~\cite{tan2020efficientdet} backbone. For the student networks, we use WDRnet with Darknet-Tiny~\cite{redmon2016you} and a lighter Darknet-Tiny-H version with half of the channels of DarkNet-tiny in each layer. Moreover, for the lighter SPNv2 student we train a $\phi$~=~0 version which is based on the EfficientDet-D0~\cite{tan2020efficientdet} backbone. A more detailed comparison of these architectures is presented in Table~\ref{tab:archs}. 

\begin{table}[h]
\centering
\renewcommand{\arraystretch}{1.3}
\setlength{\tabcolsep}{5pt}
\caption{Comparison of different architectures in terms of number of parameters (in millions), floating point operations per second (in billions), and input resolution.}
\label{tab:archs}
\resizebox{\columnwidth}{!}{%
    \begin{tabular}{clccc}
    \hline
    \textbf{Model}          & \textbf{Backbone}      & \textbf{\#Param [M]} & \textbf{\#FLOPs [B]} & \textbf{Input Res.} \\ \hline
    \multirow{3}{*}{WDRNet} & Darknet-53             & 52.1                 & 36.51           & \multirow{3}{*}{$640 \times 480$} \\ 
                            & Darknet-Tiny           & 8.5                  & 18.84           &  \\ 
                            & Darknet-Tiny-H         & 2.3                  & 4.75            &  \\ \hline
    \multirow{2}{*}{SPNv2}  & EfficientDet-D6        & 57.8                 & 288.27          & \multirow{2}{*}{$768 \times 512$} \\ 
                            & EfficientDet-D0        & 3.8                  & 12.1            &  \\ \hline
    \end{tabular}%
}
\end{table}

\noindent\textbf{Datasets.} LINEMOD~\cite{10.1007/978-3-642-37331-2_42} is a general object pose estimation benchmark. It contains 16,000 images belonging to 13 different object categories. SPEED+~\cite{park2022speed+} is a cross-domain dataset for Spacecraft pose estimation. It is formed by 60,000 synthetic images for training and two additional unlabeled  subsets containing 6,740 and 2,791 images from two different domains, \texttt{lightbox} and \texttt{sunlamp}, respectively. The \texttt{lightbox} consists of the Tango spacecraft~\cite{amico2014tango} mockup model illuminated to simulate diffuse light in the Earth’s orbit, whereas the \texttt{sunlamp} images of the same model are illuminated with a different lamp setup to simulate direct sunlight. In our experiments, we follow the same data augmentation setup used in ~\cite{park2023spnv2}.

\noindent\textbf{Metrics.} We report our results on LINEMOD using the Average Distance of Model Points (ADD) metric~\cite{10.1007/978-3-642-37331-2_42}, which is defined as the average distance between the transformed 3D model and the predicted pose. More specifically, we use the ADD-0.1d variant, which represents the percentage of images with an average distance lower than 10\% of the object diameter. For symmetric objects, the average closest point distance (ADD-S) metric~\cite{xiang2017posecnn} is used. In the experiments on SPEED+, we do not have the 3D model of the object; thus, we use the rotation, translation, and pose errors, denoted as $E_{R}$, $E_{T}$, and $E_{pose}$, respectively, as in~\cite{park2023spnv2}.

\noindent\textbf{Baselines.} We evaluate our proposed approaches against several baselines, namely, (1) a fully trained student model without KD, referred to as~\textbf{Student}, (2) the state-of-the-art KD approach called \textbf{ADLP} from~\cite{Guo_2023_CVPR}, which, to the best of our knowledge, is the only KD method specifically designed for 6DoF pose estimation, combined with the feature-based KD approach referred to as \textbf{FKD}~\cite{zhang2021improve} which has been shown to be orthogonal with ADLP. Our proposed Uncertainty-Aware KD, Prediction-related Feature KD, and their combination are labeled as \textbf{UAKD}, \textbf{PFKD}, and \textbf{UAKD+PFKD}, respectively. For UAKD and UAKD+PFKD, we report only the best average results with an ensemble based on 4 and 6 teacher models.
\subsection{6DoF Object Pose Estimation with LINEMOD} 

\noindent\textbf{Results using WDRnet.}  
We report in Table~\ref{tab:wdrnet_linemod} the results obtained for WDRnet using the DarkNet-Tiny-H and DarkNet-Tiny backbones. Our approach (UAKD + PFKD) achieves the best performance on LINEMOD for both backbones with an ADD-0.1d of 89.0 and 92.3, respectively. Remarkably, even by integrating PFKD or UAKD solely, our method outperforms the state-of-the-art in terms of ADD-0.1d on LINEMOD in both cases. In general, our method reduces the performance gap between the DarkNet53-based teacher model and the student models, namely, DarkNet-Tiny-H (which has 95.5\% fewer parameters) and DarkNet-Tiny (which has 83.7\% fewer parameters)  by 7.1 and 3.6, respectively. It can be noted that DarkNet-Tiny even surpasses the teacher model for some classes such as \textit{Bvise}, \textit{Can}, \textit{Driller}, \textit{Eggbox} and \textit{Phone}.


\begin{table*}[ht]
\centering
\captionsetup{justification=justified, singlelinecheck=false}
\caption{Performance on the LINEMOD dataset in terms of ADD-0.1d using DarkNet-Tiny-H and DarkNet-Tiny based students; Objects with * are symmetric and the ADD-S metric is used instead.}
\label{tab:wdrnet_linemod}
\begin{adjustbox}{max width=0.9\textwidth}
\begin{tabular}{cc||ccccc||ccccc}
\toprule

\multicolumn{1}{c}{\multirow{6}{*}{ {Class}}} & \multicolumn{1}{c}{\multirow{6}{*}{\makecell{Teacher \\ Model}}} & \multicolumn{5}{c}{\makecell{DarkNet-Tiny-H}} & \multicolumn{5}{c}{\makecell{DarkNet-Tiny}}  \\ \cmidrule(lr){3-7} \cmidrule(lr){8-12}
& \multicolumn{1}{c}{} & \multicolumn{2}{c}{Baselines} & \multicolumn{3}{c}{\textbf{Ours}} & \multicolumn{2}{c}{Baselines} & \multicolumn{3}{c}{\textbf{Ours}} \\ \cmidrule(lr){3-4} \cmidrule(lr){5-7} \cmidrule(lr){8-9} \cmidrule(lr){10-12}

& \multicolumn{1}{c}{}  & Student & \makecell{ADLP~\cite{Guo_2023_CVPR} \\ + FKD~\cite{zhang2021improve}} & UAKD & PFKD & \multicolumn{1}{c}{\makecell{UAKD \\ + PFKD}} & Student & \makecell{ADLP~\cite{Guo_2023_CVPR} \\ + FKD~\cite{zhang2021improve}} & UAKD & PFKD & \makecell{UAKD \\ + PFKD}  \\ \cmidrule{1-12}



\multicolumn{1}{c||}{Ape} &  83.0 &  65.4 &  69.9 &  75.6 &  77.2 &  \textbf{79.1} &  73.4 &  76.2 &  79.6 & \textbf{ 80.3} &  80.0 \\ 
\multicolumn{1}{c||}{Benchvise}  &  96.5 &  92.0 &  93.7 & \textbf{ 94.5} &  94.1 &  \textbf{94.5} &  95.2 &  96.7 & \textbf{ 97.2} &  95.3 &  96.5 \\ 
\multicolumn{1}{c||}{Camera}  &  93.8 &  78.4 &  84.5 &  88.0 & \textbf{ 89.2} &  88.0 &  91.2 &  92.0 & \textbf{ 93.5} &  92.4 &  93.4 \\ 
\multicolumn{1}{c||}{Can}  &  96.7 &  82.2 &  83.9 & \textbf{ 90.7} &  88.9 &  {89.8} &  92.4 &  94.0 &  \textbf{97.1} &  95.4 &  95.4 \\ 
\multicolumn{1}{c||}{Cat}  &  93.9 &  81.5 &  81.6 &  87.6 &  85.1 & \textbf{ 88.7} &  87.2 &  88.6 &  91.6 & \textbf{ 93.7} &  92.8 \\ 
\multicolumn{1}{c||}{Driller}  &  95.5 &  85.5 &  90.3 &  93.3 & \textbf{ 95.5} &  93.4 &  92.2 &  94.8 &  96.9 &  94.0 &  \textbf{97.0} \\ 
\multicolumn{1}{c||}{Duck}  &  79.0 &  64.3 &  68.9 & \textbf{ 72.7} &  68.7 & \textbf{ 72.7} &  70.9 &  74.7 &  \textbf{77.9} & \textbf{ 77.9} &  \textbf{77.9} \\ 
\multicolumn{1}{c||}{Eggbox\textsuperscript{*}}  &  99.2 &  95.8 &  96.4 &  97.7 &  \textbf{98.3} &  97.9 &  99.3 &  99.3 & \textbf{ 99.5} &  98.8 &  98.9 \\ 
\multicolumn{1}{c||}{Glue\textsuperscript{*}}  &  97.9 &  90.7 &  93.2 &  94.8 &  95.4 &  \textbf{96.4} &  97.2 &  97.7 &  98.0 &  98.0 &  \textbf{98.3} \\ 
\multicolumn{1}{c||}{Holepuncher}  &  87.7 &  73.2 &  76.3 &  80.1 &  \textbf{81.4} &  \textbf{81.4} &  78.0 &  82.2 &  85.5 & \textbf{ 87.1} &  85.5 \\ 
\multicolumn{1}{c||}{Iron}  &  95.5 &  86.3 &  90.5 &  89.6 &  \textbf{93.2} &  92.4 &  92.1 &  93.2 &  94.2 &  94.6 & \textbf{ 95.3} \\ 
\multicolumn{1}{c||}{Lamp}  &  98.1 &  93.6 &  94.6 &  96.4 & \textbf{ 96.5} &  96.0 &  96.6 &  96.8 &  97.8 &  97.5 &  \textbf{99.1} \\ 
\multicolumn{1}{c||}{Phone} &  91.3 &  76.0 &  79.2 &  85.0 &  83.3 & \textbf{ 86.9} &  87.5 &  89.6 &  89.8 & \textbf{ 92.9} &  89.6 \\ \cmidrule(lr){1-1} \cmidrule(lr){2-2} \cmidrule(lr){3-7} \cmidrule(lr){8-12}
\multicolumn{1}{c||}{\multirow{2}{*}{AVG.}} & \multirow{2}{*}{ 92.9} & \multirow{2}{*}{ 81.9} &  84.8 & \textbf{ 88.1} & \textbf{ 88.2} & \textbf{ 89.0} & \multirow{2}{*}{ 88.7} &  90.4 & \textbf{ 92.2} & \textbf{ 92.0} & \textbf{ 92.3} \\ \multicolumn{1}{c||}{\multirow{2}{*}{}}
  &  &  & (↑  2.9) & \textbf{(↑  6.2)} & \textbf{(↑  6.3)} & \textbf{(↑  7.1)} &  & (↑  1.7) & \textbf{(↑  3.5)} & \textbf{(↑  3.3)} & \textbf{(↑  3.6)}\\ \bottomrule
\end{tabular}
\end{adjustbox}
\vspace{-3mm}
\end{table*}


\noindent\textbf{Results using SPNv2.} Table~\ref{tab:spnv2_linemod} reports the results obtained when SPNv2~\cite{park2023spnv2} on LINEMOD.  As for WDRnet, both UAKD and PFKD, even when integrated alone, outperforms the state-of-the-art. Specifically, it  improves the results of the student model by 1.4 and 1.2 points on average LINEMOD, respectively. Moreover, by combining UAKD and PFKD, our approach (UAKD+PFKD) achieves an improvement of 1.8, surpassing the fully-trained teacher model on the \textit{Can} and \textit{Iron} classes while reducing the FLOPs by 96.6\%.
\begin{table}[h]
\centering
\caption{Performance on the LINEMOD dataset in terms of ADD-0.1d and ADD-S using an SPNv2 $\phi$ = 0 student.}
\label{tab:spnv2_linemod}
\begin{adjustbox}{max width=0.475\textwidth}
\begin{tabular}{cc||ccccc}
\toprule

\multicolumn{1}{c}{\multirow{6}{*}{ {Class}}} & \multicolumn{1}{c}{\multirow{6}{*}{\makecell{Teacher \\ Model}}} & \multicolumn{5}{c}{\makecell{SPNv2 $\phi$ = 0}} \\ \cmidrule(lr){3-7} 
& \multicolumn{1}{c}{} & \multicolumn{2}{c}{Baselines} & \multicolumn{3}{c}{\textbf{Ours}} \\ \cmidrule(lr){3-4} \cmidrule(lr){5-7}

& \multicolumn{1}{c}{} & Student & \makecell{ADLP~\cite{Guo_2023_CVPR} \\ + FKD~\cite{zhang2021improve}} & UAKD & PFKD & \makecell{UAKD \\ + PFKD} \\ \cmidrule{1-7}


\multicolumn{1}{c||}{Ape} &  {88.9} &  {67.6} &  {70.8} & \textbf{ {70.8}} & {68.8} &  {70.6} \\ 
\multicolumn{1}{c||}{Benchvise} &  {98.5} &  {91.9} &  {92.3} & \textbf{ {93.5}} &  {92.9} &  {92.8} \\ 
\multicolumn{1}{c||}{Camera} &  {96.9} &  {90.3} &  {92.4} &  {92.5} &  {92.8} &  {\textbf{93.0}} \\ 
\multicolumn{1}{c||}{Can} &  {89.7} &  {88.6} &  {88.9} &  {89.2} &  {88.2} &  {\textbf{89.8}} \\ 
\multicolumn{1}{c||}{Cat} &  {93.7} &  {90.3} &  {90.9} &  {89.6} & \textbf{ {91.0}} & \textbf{ {91.0}} \\ 
\multicolumn{1}{c||}{Driller} &  {97.1} &  {82.2} &  {82.9} &  {84.7} &  {83.2} &  {\textbf{84.9}} \\ 
\multicolumn{1}{c||}{Duck} &  {77.7} &  {68.9} &  {69.7} &  {69.0} &  {69.1} & \textbf{ {70.0}} \\ 
\multicolumn{1}{c||}{Eggbox\textsuperscript{*}} &  {95.9} &  {93.5} &  {93.6} &  {93.8} & \textbf{ {94.3}} &  {94.2} \\ 
\multicolumn{1}{c||}{Glue\textsuperscript{*}} &  {96.4} &  {92.4} &  {92.9} &  {92.9} &  {\textbf{93.0}} & \textbf{ {93.0}} \\ 
\multicolumn{1}{c||}{Holepuncher} &  {87.4} &  {75.2} &  {78.4} &  {79.3} &  {79.8} &  {\textbf{80.0}} \\ 
\multicolumn{1}{c||}{Iron} &  {86.4} &  {87.0} &  {88.0} &  {88.3} &  {88.2} &  {\textbf{88.7}} \\ 
\multicolumn{1}{c||}{Lamp} &  {95.7} &  {92.4} &  {92.7} &  {92.8} & \textbf{ {93.2}} &  {92.9} \\ 
\multicolumn{1}{c||}{Phone} &  {91.8} &  {82.2} &  {83.2} &  {83.8} &  {83.6} & \textbf{ {84.0}} \\ 
\midrule
\multicolumn{1}{c||}{\multirow{2}{*}{AVG.}} & \multirow{2}{*}{ {92.8}} & \multirow{2}{*}{ {84.8}} &  {85.9} & \textbf{ {86.2}} &  {\textbf{86.0}} &  {\textbf{86.6}} \\ 
\multicolumn{1}{c||}{\multirow{2}{*}{}} &  & &  {(↑ 1.1)} & \textbf{ {(↑ 1.4)}} &  {\textbf{(↑ 1.2)}} &  {\textbf{(↑ 1.8)}}  \\ \bottomrule
\end{tabular}%
\end{adjustbox}
\vspace{-5mm}
\end{table}

\begin{table*}[h]
\centering
\captionsetup{justification=justified, singlelinecheck=false}
\caption{Performance on the SPEED+ dataset in terms of the rotation error $E_{R}$, translation error $E_{T}$, and pose error $E_{pose}$ using an SPNv2 $\phi =0$ student under the \texttt{synthetic}, \texttt{lightbox} and \texttt{sunlamp} domains; $E^*_{pose}$ refer to the HIL pose error~\cite{park2023spnv2}.}
\label{tab:spnv2_speed+}
\begin{adjustbox}{max width=0.8\textwidth}
\begin{tabular}{ccc|ccccccccc}
\toprule

\multicolumn{3}{c}{\multirow{2}{*}{Model}} & \multicolumn{3}{c}{\texttt{synthetic}} & \multicolumn{3}{c}{\texttt{lightbox}} & \multicolumn{3}{c}{\texttt{sunlamp}} \\ \cmidrule(lr){4-6} \cmidrule(lr){7-9} \cmidrule(lr){10-12}

& & & $E_T$ [m] & $E_R$ [$^{\circ}$] & $E_{pose}$ [-] & $E_T$ [m] & $E_R$ [$^{\circ}$] & $E^*_{pose}$ [-] & $E_T$ [m] & $E_R$ [$^{\circ}$] & $E^*_{pose}$ [-] \\
\midrule
\multicolumn{3}{c|}{\multirow{1}{*}{Teacher Model}} &  0.042 &  1.010 &  0.024 &  0.272 &  8.968 &  0.202 &  0.259 &  12.580 &  0.263 \\

\midrule

\multicolumn{1}{c|}{\multirow{5}{*}{\rotatebox{90}{SPNv2 $\phi$ = 0}}}  & \multicolumn{1}{c|}{\multirow{2}{*}{\makecell{\rotatebox{90}{Base.}}}} & Student &  0.050 &  1.441 &  0.033 &  0.447 &  16.804 &  0.368 &  0.372 &  19.366 &  0.401 \\

\multicolumn{1}{c}{\multirow{1}{*}{\makecell{}}} \vrule & \multicolumn{1}{c}{\multirow{1}{*}{\makecell{}}}  \vrule & ADLP\cite{Guo_2023_CVPR}+ FKD~\cite{zhang2021improve} &  0.045 &  1.157 &  0.027 &  0.482 &  14.596 &  0.336 &  0.387 &  18.477 &  0.388 \\ 

\cmidrule(lr){3-12}
 
\multicolumn{1}{c}{\multirow{1}{*}{\makecell{}}} \vrule & \multicolumn{1}{c|}{\multirow{3}{*}{\makecell{\rotatebox{90}{\textbf{Ours}}}}}  & UAKD &  \textbf{0.041}  & 1.018 &  0.024 &  0.346 &  13.195 &  0.288 &  0.383 &  17.039 &  0.362 \\

\multicolumn{1}{c}{\multirow{1}{*}{\makecell{}}} \vrule & \multicolumn{1}{c}{\multirow{1}{*}{\makecell{}}} \vrule & PFKD &  0.048 &  1.197 &  0.028 &  0.288 &  11.973 &  0.257 &  0.388 &  17.093 &  0.364 \\

\multicolumn{1}{c}{\multirow{1}{*}{\makecell{}}} \vrule & \multicolumn{1}{c}{\multirow{1}{*}{\makecell{}}} \vrule & UAKD + PFKD  & 0.042 & \textbf{ 1.007} & \textbf{ 0.024} &  \textbf{0.288} &  \textbf{11.419} &  \textbf{0.248} &  \textbf{0.373} &  \textbf{17.026} &  \textbf{0.360} \\
 
\bottomrule
\end{tabular}%
\end{adjustbox}
\vspace{-5mm}
\end{table*}

\subsection{Spacecraft Pose Estimation with SPEED+}
For the SPEED+ \cite{park2022speed+} dataset, we conduct experiments using SPNv2 only, given the low performance of WDRnet on this specific dataset.  Each trained model is evaluated across the three different domains provided by SPEED+.
Table~\ref{tab:spnv2_speed+} provides the results obtained on SPEED+ in terms of translation, rotation and pose errors for the three different domains, namely, \texttt{synthetic}, \texttt{lightbox}, and \texttt{sunlamp}. Both the proposed UAKD and PFKD outperform the state-of-the-art, i.e., ADLP+FKD. Specifically, Our UAKD method achieves a pose error of 0.024 in the \texttt{synthetic} domain and 0.288 in the \texttt{lightbox} domain, competing with the fully trained SPNv2 model ($\phi=6$) on the \texttt{synthetic} domain and surpassing the student model by 0.080 in the \texttt{lightbox} domain. In the \texttt{sunlamp} domain, UAKD reduces the pose error by 0.039, corresponding to a decrease of 2.327$^{\circ}$ in terms of rotation error compared to the student and 1.438$^{\circ}$ compared to ADLP+FKD. Additionally, our PFKD enhances the performance, slightly improving the pose error by 0.001, 0.079, and 0.024, in the \texttt{synthetic}, \texttt{lightbox}, and \texttt{sunlamp} domains, respectively as compared to ADLP+FKD. With our UAKD+PFKD variant, the performance of the student model aligns with the teacher in the \texttt{synthetic} domain, offering a 0.003$^{\circ}$ advantage in terms of rotation error, while significantly improving the performance in the \texttt{lightbox} and \texttt{sunlamp} domains by reducing the pose error by 0.120 and 0.041, respectively, corresponding to  a reduction of 5.385$^{\circ}$ and 2.340$^{\circ}$ in terms of rotation error.


\subsection{Additional Analysis}

\begin{figure}[h]
    \centering
    \includegraphics[width=0.45\textwidth]{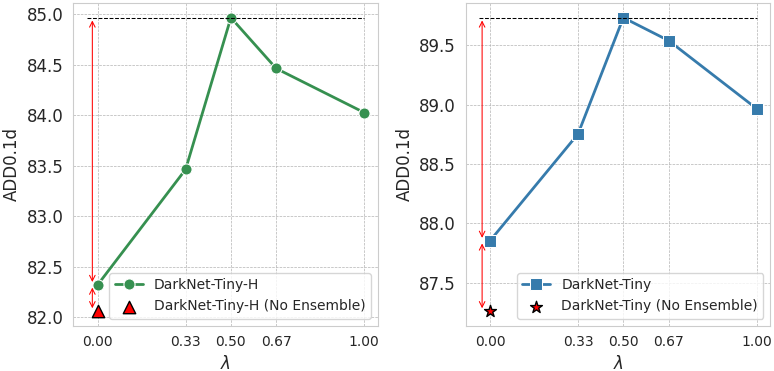}
    \caption{Impact of $\lambda$ on the ADD-0.1d Metric using WDRnet on LINEMOD.}
    \label{fig:lambda}
    \vspace{-8mm}
\end{figure}

\noindent\textbf{Effect of the Ensemble.} We examine the impact of using the ensemble strategy in our UAKD, as illustrated in Figure~\ref{fig:lambda}. We compare the results obtained for a student backbone based on an ensemble of teachers without considering uncertainty ($\lambda = 0$) to the same backbone without relying on an ensemble. We find that the ensemble contributes with an improvement of 0.2  on DarkNet-Tiny-H and 0.5 on  DarkNet-Tiny. This indicates that the increase in performance obtained by our approach returns mostly to the uncertainty-aware knowledge distillation rather than the ensembing approach. The slight improvement coming from the ensemble is attributed to its tendency to produce more accurate predictions~\cite{fort2019deep}.

\noindent\textbf{Choice of $\lambda$.} As presented in~(\ref{eq:lambda}), we introduce a weight factor $\lambda$ to balance between the WDRNet's existence score per keypoint and the uncertainty value generated by our ensemble. Figure \ref{fig:lambda} illustrates the ADD-0.1d values across various LINEMOD training sessions for different $\lambda$ values. A 50\%-50\% combination ($\lambda = 0.5$) of uncertainty and existence scores yields the best results on LINEMOD.

\section{Conclusion}
\label{sec:conclusion}

In this paper, we present an uncertainty-aware end-to-end knowledge distillation (KD) framework for efficient two-stage 6DoF pose estimation. We argue that the varying uncertainty in teacher predictions is a valuable asset for KD and propose an uncertainty-driven selective weighted distillation method, UAKD, to optimize the alignment between teacher and student keypoints. Leveraging a deep ensemble strategy, we quantify uncertainties for \textit{prediction-level} distillation and extend this to \textit{feature-level} distillation (PFKD) by tracing keypoint predictions to associated feature map regions, enabling a unified uncertainty-aware distillation. Experimental results demonstrate the effectiveness of the proposed method in terms of both accuracy and compactness, highlighting its potential for deployment in pose estimation applications. Future directions include integrating an uncertainty-guided PnP approach~\cite{Peng_2019_CVPR} and exploring other compression techniques to further improve efficiency and accuracy.

\vspace{-2mm}



\section*{Acknowledgement} 
The present work is supported by the National Research Fund (FNR), Luxembourg, under the C21/IS/15965298/ELITE project, and by Infinite Orbits. Experiments were performed on the Luxembourg national supercomputer MeluXina at LuxProvide.

\bibliographystyle{IEEEtran}

\bibliography{IEEEabrv.bib, IEEEfull.bib}
\addtolength{\textheight}{-12cm}   

\end{document}